\documentclass{amsart}

\usepackage{amsmath}
\usepackage{amsfonts}
\usepackage{latexsym}
\usepackage{graphicx}
\usepackage{multicol}
\usepackage[all]{xy}

\usepackage{mathrsfs}

\newcommand{\absolute}[1]{\lvert#1\rvert}
\newcommand{\mathset}[1]{{\{#1\}}} 
\DeclareMathOperator{\PGL}{PGL}
\DeclareMathOperator{\dist}{dist}

\newtheorem{thm}{Theorem}[section]

\newtheorem{lem}[thm]{Lemma}

\newtheorem{rem}[thm]{Remark}
\newtheorem{dfn}[thm]{Definition}

\newtheorem{exa}[thm]{Example}

\begin{document}

\title{Families of dendrograms }

\author{Patrick Erik Bradley}

\maketitle

\begin{abstract}
A conceptual framework for cluster analysis from the viewpoint of $p$-adic
geometry is introduced by describing the space of all dendrograms
for $n$ datapoints and relating it to the moduli space of 
$p$-adic Riemannian spheres with punctures using a method recently applied by
Murtagh (2004b). This method embeds a dendrogram as a subtree into the Bruhat-Tits tree
associated to the $p$-adic numbers, and goes back to
Cornelissen et al.\ (2001)
in $p$-adic geometry.
After explaining the definitions, the  concept of classifiers is
discussed in the context of moduli spaces, and upper bounds for the number
of hidden vertices in dendrograms are given. 
\end{abstract}

\section{Introduction}

Dendrograms are ultrametric spaces, and  ultrametricity is a pervasive property
of observational data, and by Murtagh (2004a) this offers computational advantages
 and a well understood basis for developping data processing tools
originating in $p$-adic arithmetic. The aim of this article
is to show that the foundations can be laid much deeper by taking into account
a natural object in $p$-adic geometry, namely the {\em Bruhat-Tits tree}.
This locally finite, regular tree 
naturally contains the dendrograms as subtrees which are 
uniquely determined by 
assigning  $p$-adic numbers to data.
Hence, the classification task is conceptionally 
reduced to finding a suitable $p$-adic data encoding.
Dragovich and Dragovich (2006) 
find a $5$-adic encoding of  
%represent 
DNA-sequences, 
%$5$-adically, 
and 
Bradley (2007) shows that strings 
have natural $p$-adic encodings.

The geometric approach makes it possible to  treat time-dependent data
on an equal footing as data that relate only to  one instant of time by providing
the concept of {\em family of dendrograms}.
Probability distributions on families are then seen as a
convenient way of describing classifiers.

Our illustrative  toy data set for this article is given as follows:

\begin{exa}
Consider the data set $D=\mathset{0,1,3,4,12,20,32,64}$
 given by $n=8$ natural numbers. We want to hierarchically classify
it with respect to the $2$-adic norm $\absolute{\cdot}_2$ as our distance function, as defined in Section \ref{p-adgeo}. 
\end{exa}

\section{A brief introduction to $p$-adic geometry} \label{p-adgeo}

Euclidean geometry is modelled on the field $\mathbb{R}$ of real numbers
which are often represented as decimals, i.e.\ expanded in powers 
of the number $10^{-1}$:
$$
x=\sum\limits_{\nu=m}^\infty a_\nu 10^{-\nu},\quad a_\nu\in\mathset{0,\dots,9},
\quad m\in\mathbb{Z}.
$$
 In this way, $\mathbb{R}$ completes the field
$\mathbb{Q}$ of rational numbers with respect to the absolute norm
$
\absolute{x}=\begin{cases}x,&x\ge 0\\-x,&x<0\end{cases}.
$
On the other hand, the $p$-adic norm on $\mathbb{Q}$ with
$$
\absolute{x}_p=\begin{cases}p^{-\nu_p(x)},&x\neq 0\\0,&x=0\end{cases}
$$
is defined for $x=\frac{a_1}{a_2}$
 by the difference  $\nu_p(x)=\nu_p(a_1)-\nu_p(a_2)\in\mathbb{Z}$ 
in the multiplicities with which
numerator and denominator of $x$  are
divisible by the prime number $p$:
$a_i=p^{\nu_p(a_i)}u_i$, and $u_i$ not divisible by $p$, $i=1,2$.

The $p$-adic norm satisfies the {\em ultrametric triangle inequality}
$$
\absolute{x+y}_p\le\max\mathset{\absolute{x}_p,\absolute{y}_p}.
$$
Completing $\mathbb{Q}$ with respect to the $p$-adic norm yields
the field $\mathbb{Q}_p$ of {\em $p$-adic numbers} which is well known to 
consist of the power series
\begin{align}
x=\sum\limits_{\nu=m}^\infty a_\nu p^{\nu},\quad a_\nu\in\mathset{0,\dots,p-1},\quad m\in\mathbb{Z}.
\label{power-p}
\end{align}
Note, that the $p$-adic expansion is in increasing powers of $p$, whereas in the decimal expansion, it is the powers of $10^{-1}$ which increase arbitrarily.
An introduction to $p$-adic numbers is e.g.\ Gouv\^ea (2003).

\begin{exa}
For our toy data set $D$, we have
$\absolute{0}_2=0$, $\absolute{1}_2=\absolute{3}_2=1$, 
$\absolute{4}_2=\absolute{12}_2=\absolute{20}_2=2^{-2}$,
$\absolute{32}_2=2^{-5}$, $\absolute{64}_2=2^{-6}$, i.e.\ $\absolute{\cdot}_2$ is maximally $1$ on $D$. Other examples: $\absolute{3/2}_3=\absolute{6/4}_3=3^{-1}$, $\absolute{20}_5=5^{-1}$, $\absolute{p^{-1}}_p=\absolute{p}_p^{-1}=p$.
\end{exa}

Consider the unit disk
$
\mathbb{D}=\mathset{x\in\mathbb{Q}_p\mid\absolute{x}_p\le 1}=B_1(0).
$ 
It consists of the so-called {\em $p$-adic integers}, and is often
denoted as $\mathbb{Z}_p$ when emphasizing its ring structure, i.e.\  closedness under 
addition, subtraction and multiplication. 
A $p$-adic number $x$ lies in an arbitrary closed disk
$
B_{p^{-r}}(a)=\mathset{x\in\mathbb{Q}_p\mid \absolute{x-a}_p\le p^{-r}}$,
where $r\in\mathbb{Z}$,
 if and only if $x-a$ is divisible by $p^r$. This condition is
equivalent to $x$ and $a$ having the first $r$ terms in common in their
$p$-adic expansions (\ref{power-p}).
The possible radii are all integer powers of $p$,
so the disjoint disks 
$
B_{p^{-1}}(0), B_{p^{-1}}(1),\dots,B_{p^{-1}}(p-1)
$
are the maximal proper subdisks of $\mathbb{D}$, as
they correspond to truncating the power series (\ref{power-p})
after the constant term. There is a unique minimal disk in which
$\mathbb{D}$ is contained properly, namely
$
B_p(0)=\mathset{x\in\mathbb{Q}_p\mid \absolute{x}_p\le p}.
$
These observations hold true for  
arbitrary $p$-adic disks, i.e.\ any disk $B_{p^{-r}}(x)$, $x\in\mathbb{Q}_p$,
 is partitioned into precisely $p$ maximal subdisks
and lies properly in a unique minimal disk. Therefore, if we define a graph
$\mathscr{T}_{\mathbb{Q}_p}$ whose
vertices are the $p$-adic disks, and edges are given by minimal inclusion,
then  every vertex of $\mathscr{T}_{\mathbb{Q}_p}$ has
precisely $p+1$ outgoing edges. In other words,
$\mathscr{T}_{\mathbb{Q}_p}$ is a {\em $p+1$-regular} tree,
and $p$ is the size of the residue field $\mathbb{F}_p=\mathbb{Z}_p/p\mathbb{Z}_p$.

\begin{dfn}\rm \label{BTT}
The tree $\mathscr{T}_{\mathbb{Q}_p}$ is called the {\em Bruhat-Tits tree}
for $\mathbb{Q}_p$.
\end{dfn}

\begin{rem}
Definition \ref{BTT} is not the usual way to define 
$\mathscr{T}_{\mathbb{Q}_p}$. The problem with this ad-hoc definition is that it does not allow for any action
of the projective linear group $\PGL_2(\mathbb{Q}_p)$. A definition invariant under projective linear transformations can be found e.g.\ in Herrlich (1980) or Bradley (2006).
\end{rem}

An important observation is that any infinite descending chain
\begin{align}
B_1\supseteq B_2\supseteq \dots \label{halfline}
\end{align}
of strictly decreasing $p$-adic disks converges to a unique $p$-adic number
$
\mathset{x}=\bigcap\limits_n B_n.
$
A chain (\ref{halfline}) defines a halfline in the Bruhat-Tits tree $\mathscr{T}_{\mathbb{Q}_p}$. Halflines differing only by finitely many vertices are said to be {\em equivalent}, and the equivalence classes under this equivalence relation are called {\em ends}. Hence the observation means that the $p$-adic numbers correspond to ends of $\mathscr{T}_{\mathbb{Q}_p}$. There is a unique end
$
B_1\subseteq B_2\subseteq\dots
$
coming from any strictly increasing sequence of disks. This end corresponds to the point at infinity in the $p$-adic projective line
$
\mathbb{P}^1(\mathbb{Q}_p)=\mathbb{Q}_p\cup\mathset{\infty},
$
whence the well known fact:

\begin{lem} \label{ends}
The ends of $\mathscr{T}_{\mathbb{Q}_p}$ are in one-to-one correspondance
with the $\mathbb{Q}_p$-rational points  of the $p$-adic projective line $\mathbb{P}^1$, i.e.\ with the elements of $\mathbb{P}^1(\mathbb{Q}_p)$.
\end{lem}

From the viewpoint of geometry, it is important to distinguish between the
$p$-adic projective line $\mathbb{P}^1$ as a $p$-adic manifold and its set 
$\mathbb{P}^1(\mathbb{Q}_p)$ 
of $\mathbb{Q}_p$-rational points, in the same way as one distinguishes between the affine real line $\mathbb{A}^1$ as a real manifold and its rational points $\mathbb{A}^1(\mathbb{Q})=\mathbb{Q}$, for example. 
One reason for  distinguishing between a space and its points is:

\begin{lem}
Endowed with the metric topology from $\absolute{\cdot}_p$,
the topological space $\mathbb{Q}_p$ is totally disconnected.
\end{lem}

The usual approaches towards defining more useful topologies on $p$-adic spaces are by introducing more points. Such an approach is the {\em Berkovich topology}, which we will very briefly describe.
More details can be found in Berkovich (1990).

The idea is to allow disks whose radii are  arbitrary positive real numbers,
not merely powers of $p$ as before.
Any strictly descending chain of such disks gives a point in the
sense of Berkovich. For the 
%Berkovich 
$p$-adic line $\mathbb{P}^1$ this amounts to:

\begin{thm}[Berkovich]
$\mathbb{P}^1$ is non-empty, compact, hausdorff and arc-wise connected.
Every point of $\mathbb{P}^1\setminus\mathset{\infty}$ corresponds to a descending sequence $B_1\supseteq B_2\supseteq \dots$ of $p$-adic disks such that $B=\bigcap B_n$ is one of the following:
\begin{enumerate}
\item a point $x$ in $\mathbb{Q}_p$,
\item a closed $p$-adic disk with radius $r\in\absolute{\mathbb{Q}_p}_p$,
\item a closed $p$-adic disk with radius $r\notin\absolute{\mathbb{Q}_p}_p$,
\item empty.
\end{enumerate} 
\end{thm}

Points of types 2.\ to 4.\ are called {\em generic}, points of type 1.\
{\em classical}. 
We remark that Berkovich's definition of points is technically somewhat different  and allows to define more general $p$-adic spaces.
Finally, the Bruhat-Tits tree $\mathscr{T}_{\mathbb{Q}_p}$ is recovered inside
$\mathbb{P}^1$:

\begin{thm}[Berkovich] \label{retract}
$\mathscr{T}_{\mathbb{Q}_p}$ is a  retract of $\mathbb{P}^1\setminus\mathbb{P}^1(\mathbb{Q}_p)$, i.e.\ there is a map 
$\mathbb{P}^1\setminus\mathbb{P}^1(\mathbb{Q}_p)\to\mathscr{T}_{\mathbb{Q}_p}$
whose restriction to $\mathscr{T}_{\mathbb{Q}_p}$ is the identity map on $\mathscr{T}_{\mathbb{Q}_p}$.
\end{thm}

%Theorem \ref{retract} is not the only way to relate $\mathscr{T}_{\mathbb{Q}_p}$ to $\mathbb{P}^1$, but sufficient for us.

\section{$p$-adic dendrograms}
\begin{figure}[h]
%\begin{minipage}[b]{.4\textwidth}
%\centering
$$
\begin{array}{c|cccccccc}
\nu_2&0&1&3&4&12&20&32&64\\ \hline
0&\infty&0&0&2&2&2&5&6\\
 1&0&\infty&1&0&0&0&0&0\\
 3&0&1&\infty&0&0&0&0&0\\
 4&2&0&0&\infty&3&4&2&2\\
12&2&0&0&3&\infty&3&2&2\\
20&2&0&0&4&3&\infty&2&1\\
32&5&0&0&2&2&2&\infty&5\\
64&6&0&0&2&2&1&5&\infty
\end{array}
$$
%
%\vspace*{7mm}
\caption{$2$-adic valuations for $D$.}
\label{dist2}
\end{figure}

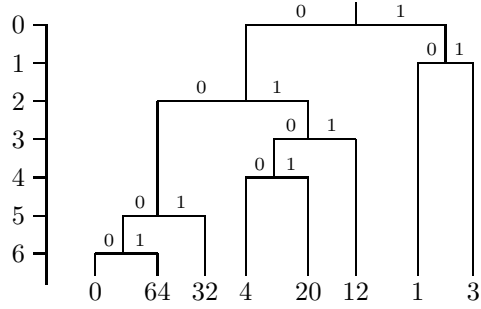
\begin{figure}[h]
%\end{minipage}
%\hfill
%\begin{minipage}[b]{.55\textwidth}
%\centering
%\hspace*{-7mm}
%\includegraphics[scale=.23]{dendro.eps}
$$
\xymatrix@R=2pt@C=2pt{
&&&&&&&&&&\ar@{-}[d]&&&&\\
0\ar@{-}[r]&*\txt{}\ar@{-}[ddddddd]&&&&&&*\txt{}\ar@{-}[dd]&&&*\txt{}\ar@{-}[rrr]^1\ar@{-}[lll]_0&&&*\txt{}\ar@{-}[d]&
\\
1\ar@{-}[r]&*\txt{}&&&&&&&&&&&*\txt{}\ar@{-}[dddddd]&*\txt{}\ar@{-}[l]_0\ar@{-}[r]^1&*\txt{}\ar@{-}[dddddd]
\\
2\ar@{-}[r]&*\txt{}&&&&*\txt{}\ar@{-}[ddd]&&*\txt{}\ar@{-}[ll]_0\ar@{-}[rr]^1&&*\txt{}\ar@{-}[d]&&&&&
\\
3\ar@{-}[r]&*\txt{}&&&&&&&*\txt{}\ar@{-}[d]&*\txt{}\ar@{-}[l]_0\ar@{-}[r]^1&*\txt{}\ar@{-}[dddd]&&&&
\\
4\ar@{-}[r]&*\txt{}&&&&&&*\txt{}\ar@{-}[ddd]&*\txt{}\ar@{-}[l]_0\ar@{-}[r]^1&*\txt{}\ar@{-}[ddd]&&&&&
\\
5\ar@{-}[r]&*\txt{}&&&*\txt{}\ar@{-}[d]&*\txt{}\ar@{-}[l]_0\ar@{-}[r]^1&*\txt{}\ar@{-}[dd]&&&&&&&&
\\
6\ar@{-}[r]&*\txt{}&&*\txt{}\ar@{-}[d]&*\txt{}\ar@{-}[l]_0\ar@{-}[r]^1&*\txt{}\ar@{-}[d]&&&&&&&&&
\\
&&&0&&64&32&4&&20&12&&1&&3
\\
}
$$
\caption{$2$-adic dendrogram for $D\cup\mathset{\infty}$.}
\label{Murtaghdendro}
%\end{minipage}
\end{figure}

\begin{exa}
The  $2$-adic distances within $D$ are encoded in Figure \ref{dist2},
where $\dist(i,j)=2^{-\nu_2(i,j)}$, if $\nu_2(i,j)$
 is the corresponding
entry in Figure \ref{dist2}, 
using $2^{-\infty}=0$.  Figure \ref{Murtaghdendro}
is the dendrogram for $D$ using $\absolute{\cdot}_2$: 
the distance between disjoint clusters equals the distances
between any of their representatives.
\end{exa}

Let $X\subseteq\mathbb{P}^1(\mathbb{Q}_p)$ be a finite set.
By Lemma \ref{ends}, a point of $X$ can be considered as an end in $\mathscr{T}_{\mathbb{Q}_p}$. 

\begin{dfn}\rm
The smallest subtree $\mathscr{D}(X)$ of $\mathscr{T}_{\mathbb{Q}_p}$ 
whose ends are given by $X$ is called the {\em $p$-adic dendrogram} 
for $X$.
\end{dfn} 

Cornelissen et al.\ (2001) use $p$-adic dendrograms for studying $p$-adic symmetries, 
cf.\ also Cornelissen and Kato (2005).
We will ignore vertices in $\mathscr{D}(X)$ from which precisely two edges emanate. Hence, for example, $\mathscr{D}(\mathset{0,1,\infty})$ consists of a unique 
vertex $v(0,1,\infty)$ and three ends.
The dendrogram for a set $X\subseteq\mathbb{N}\cup\mathset{\infty}$ containing $\mathset{0,1,\infty}$ is a rooted tree with root $v(0,1,\infty)$. 

\begin{exa}
The $2$-adic dendrogram in Figure \ref{Murtaghdendro} is nothing but $\mathscr{D}(X)$ for
$X=D\cup\mathset{\infty}$ and is in fact inspired by the first dendrogram of Murtagh (2004b). The path from the top  cluster to $x_i$ yields its binary  representation $[\cdot]_2$ which easily translates into the $2$-adic expansion:
$0=[0000000]_2$, $64=[1000000]_2=2^6$, $32=[0100000]_2=2^5$, 
$4=[0000100]_2=2^2$, $20=[0010100]_2=2^2+2^4$, $12=[0001100]_2=2^2+2^3$, 
$1=[0000001]_2$, $3=[0000011]_2=1+2^1$. 
\end{exa}

Any encoding of some data set $M$ which assigns to each $x\in M$ a
$p$-adic representation of an integer including $0$ and $1$, yields
a $p$-adic dendrogram $\mathscr{D}(M\cup\mathset{\infty})$
whose root is $v(0,1,\infty)$,
and any dendrogram for real data can be embedded
in a non-unique way into $\mathscr{T}_{\mathbb{Q}_p}$
as a $p$-adic dendrogram in such a way that $v(0,1,\infty)$ represents the top cluster,
if $p$ is large enough. In particular, any binary dendrogram is a $2$-adic
dendrogram.
However, a little algebra helps to find sufficiently large
$2$-adic Bruhat-Tits trees $\mathscr{T}_K$ which allow embeddings of 
arbitrary dendrograms into $\mathscr{T}_K$. In fact, by $K$ we mean
a finite extension field of $\mathbb{Q}_p$. The $p$-adic norm $\absolute{\cdot}_p$ extends uniquely to a norm $\absolute{\cdot}_K$ on $K$, for which it is a complete field, called a {\em $p$-adic number field}. 
The {\em integers of $K$} are again the unit disk
$\mathcal{O}_K=\mathset{x\in K\mid \absolute{x}_K\le 1}$, and the role of the prime $p$ is played by a so-called {\em uniformiser} $\pi\in\mathcal{O}_K$.
It has the property that $\mathcal{O}_K/\pi\mathcal{O}_K$ is a finite field
with $q=p^f$ elements and
contains $\mathbb{F}_p$. Hence, if some dendrogram has a vertex with maximally $n\ge 2$ children, then we need $K$ large enough such that $2^f\ge n$.
This is possible by the results of number theory. Restricting 
to the prime characteristic $2$ has not only the advantage 
of avoiding the need to switch the prime number $p$ in the case of
more than $p$ children vertices, but also the arithmetic in $2$-adic
number fields is known to be computationally simpler, especially 
as in our case the so-called {\em unramified} extensions,
i.e.\ where $\dim_{\mathbb{Q}_2}K=f$, are sufficient.

\begin{exa}
According to Bradley (2007), strings over a finite alphabet can be encoded in an unramified extension of $\mathbb{Q}_p$, and hence be classified $p$-adically.
\end{exa}

\section{The space of dendrograms}

From now on, we will formulate everything for the case $K=\mathbb{Q}_p$, bearing in mind that
all results hold true for general $p$-adic number fields $K$.
Let $S=\mathset{x_1,\dots,x_n}\subseteq\mathbb{P}^1(\mathbb{Q}_p)$ consist of $n$ distinct classical  points of $\mathbb{P}^1$ 
such that $x_1=0$, $x_2=1$, $x_3=\infty$.
Similarly as in Theorem \ref{retract}, the $p$-adic dendrogram 
$\mathscr{D}(S)$
is a  retract of the marked projective line $X=\mathbb{P}^1\setminus S$. We call $\mathscr{D}(S)$ the {\em skeleton} of $X$.
  The space of all projective lines with $n$ such markings is denoted by
$\mathfrak{M}_{n}$, and the space of corresponding $p$-adic dendrograms
by $\mathfrak{D}_{n-1}$. $\mathfrak{M}_{n}$ is a $p$-adic space of dimension $n-3$, its skeleton
 $\mathfrak{D}_{n-1}$ is a cw-complex of real polyhedra
whose cells of maximal dimension $n-3$ consist of the binary dendrograms.
Neighbouring cells are passed through by contracting bounded edges
as the $n-3$ ``free'' markings ``move'' about $\mathbb{P}^1$ without colliding.
For example, $\mathfrak{M}_3$ is just a point corresponding to $\mathbb{P}^1\setminus\mathset{0,1,\infty}$. $\mathfrak{M}_4$ has one free marking $\lambda$
which can be any $\mathbb{Q}_p$-rational point from $\mathbb{P}^1\setminus\mathset{0,1,\infty}$.
\begin{figure}[h]
$$
\begin{array}{c}
\xymatrix@=3pt{
&&\infty&&\\
 A\colon&&*\txt{$\bullet$}\ar@{-}[u]\ar@{-}[dl]\ar@{-}[ddrr]&&\\
&*\txt{$\bullet$}\ar@{-}[dl]\ar@{-}[dr]&&&\\
0&&1&&\lambda}
\end{array}
\hfill
\begin{array}{c}
\xymatrix@=3pt{
&&\infty&&\\
 B\colon&&*\txt{$\bullet$}\ar@{-}[u]\ar@{-}[dl]\ar@{-}[ddrr]&&\\
&*\txt{$\bullet$}\ar@{-}[dl]\ar@{-}[dr]&&&\\
0&&\lambda&&1}
\end{array}
\hfill
\begin{array}{c}
\xymatrix@=3pt{
&&\infty&&\\
 C\colon&&*\txt{$\bullet$}\ar@{-}[u]\ar@{-}[dl]\ar@{-}[ddrr]&&\\
&*\txt{$\bullet$}\ar@{-}[dl]\ar@{-}[dr]&&&\\
1&&\lambda&&0}
\end{array}
\hfill
\begin{array}{c}
\xymatrix@=10pt{
&\infty&\\
v\colon&*\txt{$\bullet$}\ar@{-}[u]\ar@{-}[dl]\ar@{-}[d]\ar@{-}[dr]&\\
0&1&\lambda
}
\end{array}
$$
\caption{Dendrograms representing the different regions of $\mathfrak{D}_3$.}
\label{d3}
\end{figure}
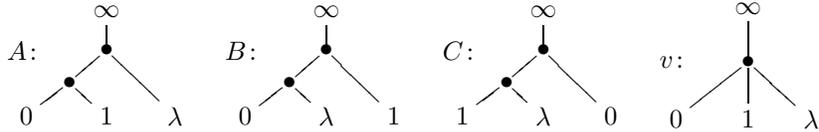
Hence, the skeleton $\mathfrak{D}_3$ is itself a binary dendrogram with precisely one vertex $v$ and three unbounded edges $A,B,C$ (cf.\ Figure \ref{d3}).
For $n\ge 3$ there are maps
$$
f_{n+1}\colon\mathfrak{M}_{n+1}\to\mathfrak{M}_n,
\quad\phi_{n+1}\colon\mathfrak{D}_{n}\to\mathfrak{D}_{n-1},
$$
which forget the $(n+1)$-st marking. Consider a $\mathbb{Q}_p$-rational point $x\in\mathfrak{M}_{n}$,
corresponding to $\mathbb{P}^1\setminus S$  with skeleton $d$.
Its fibre $f^{-1}_{n+1}(x)$ corresponds to  $\mathbb{P}^1\setminus S'$ for all possible
$S'$ whose first $n$ entries constitute $S$. Hence, the extra marking $\lambda\in S'\setminus S$ can be taken arbitrarily from $\mathbb{P}(\mathbb{Q}_p)\setminus S$. In this way, the space
 $f^{-1}_{n+1}(x)$
can be considered as $\mathbb{P}^1\setminus S$, and $\phi_{n+1}^{-1}(d)$
as the $p$-adic dendrogram for $S$.
What we have seen is that taking fibres recovers the dendrograms corresponding
to points in the space $\mathfrak{D}_n$.
Instead of  fibres of points, one can take fibres of arbitrary subspaces:

\begin{dfn}\rm
A {\em family of dendrograms with $n$ data points over a space $Y$} is a map $Y\to\mathfrak{D}_n$ from some $p$-adic space $Y$ to $\mathfrak{D}_n$.
\end{dfn}

For example, take $Y=\mathset{y_1,\dots,y_T}$.
Then a family $Y\to\mathfrak{D}_n$  is a time series of $n$ collision-free 
particles, if $t\in\mathset{1,\dots,T}$ is interpreted as time variable. 
It is also possible to take into account colliding particles by using
compactifications of $\mathfrak{M}_n$  as described in Bradley (2006).

\section{Distributions on dendrograms}

Given a dendrogram $\mathscr{D}$ for some data  $S=\mathset{x_1,\dots,x_n}$,
the idea of a classifier is to incorporate a further datum $x\notin S$
into the classification scheme represented by $\mathscr{D}$.
Often this is done by assigning probabilities to the vertices of $\mathscr{D}$,
 depending on $x$. The result is then a family of possible
dendrograms for $S\cup\mathset{x}$ with a certain probability distribution.
It is clear that, in the case of $p$-adic dendrograms, this family is nothing but
$\phi^{-1}_{n+1}(d)\to\mathfrak{D}_n$, if $d\in\mathfrak{D}_{n-1}$ is the point 
representing $\mathscr{D}$.
This motivates the following definition:

\begin{dfn}\rm
A {\em universal $p$-adic classifier $\mathcal{C}$ for $n$ given points} is a probability
distribution on $\mathfrak{M}_{n+1}$.
\end{dfn}

Here, we take on $\mathfrak{M}_{n+1}$ the Borel $\sigma$-algebra  associated
to the open sets of the Berkovich topology. If $x\in\mathfrak{M}_{n}$
corresponds to  $\mathbb{P}^1\setminus S$, then $\mathcal{C}$
induces a distribution on $f^{-1}_{n+1}(x)$, hence (after renormalisation)
a probability distribution on $\phi^{-1}_{n+1}(d)$, where
$d\in\mathfrak{D}_{n-1}$ is the point corresponding to the dendrogram $\mathscr{D}(S)$. The similar holds true for general families of dendrograms, e.g.\ time series of particles.

\section{Hidden vertices}

A vertex $v$ in a $p$-adic dendrogram $\mathscr{D}$ is called {\em hidden},
if 
 the class
corresponding to $v$
 is not the top class and 
does not directly contain data points but
is composed of non-trivial subclasses.
The subforest of $\mathscr{D}$ spanned by its hidden vertices will be
denoted by $\mathscr{D}^h$, and is called the {\em hidden part} of 
$\mathscr{D}$. The number $b_0^h$ of connected components of $\mathscr{D}^h$ 
measures how the clusters corresponding to non-hidden vertices are spread 
within
the dendrogram $\mathscr{D}$. We give bounds for $b_0^h$ and 
 the number $v^h$ of hidden vertices, and refer to Bradley (2006) for the combinatorial
proofs (Theorems 8.3 and 8.5).

\begin{thm}
Let $\mathscr{D}\in\mathfrak{D}_n$. Then
$$
v^h\le\frac{n+1}{4}-b_0^h+1\quad\text{and}\quad
b^h_0\le\frac{n-4}{3}, 
$$
where the latter bound  is sharp.
\end{thm}

\section{Conclusions}

Since ultrametricity is the natural property which allows 
classification  and is pervasive in observational data,
the techniques of ultrametric analysis and 
$p$-adic geometry
are at ones disposal for identifying and exploiting ultrametricity.
A $p$-adic encoding of data provides a 
 way to investigate arithmetic
properties of the $p$-adic numbers representing the data. 

It is our aim to lay the  geometric foundation towards
$p$-adic data encoding. From the geometric point of view
it is natural to perform the encoding by embedding its underlying
dendrogram into the Bruhat-Tits tree. In fact, the dendrogram and
its embedding are
uniquely determined by the $p$-adic numbers  representing the data.
For this end, we give an account of $p$-adic geometry  in order to define
$p$-adic dendrograms as subtrees of the Bruhat-Tits tree.

In the next step we introduce  the space of all dendrograms
for a given number of data points 
which, by $p$-adic geometry, is contained in the  space
$\mathfrak{M}_n$ of all marked projective lines, an object appearing
in the context of the classification of Riemann surfaces.
The advantages of considering the space of dendrograms 
rely on the fact
that a conceptual formulation of moving particles as 
families of dendrograms is made possible, and its simple geometry
as a polyhedral complex. Also, assigning distributions on $\mathfrak{M}_n$
allows for probabilistic incorporation of further data to 
a given
dendrogram.
At the end, we give bounds for the numbers of hidden vertices and
hidden components
of dendrograms.

What remains to do is to computationally exploit the foundations laid
in this article by 
developping a code along these lines 
and apply it to Fionn Murtagh's task of finding ultrametricity in data.

\section*{Acknowledgements}
The author is supported by the Deutsche Forschungsgemeinschaft through
the research project {\em Dynamische Geb\"audebestandsklassifikation}
BR 3513/1-1, and thanks 
 Hans-Hermann Bock for suggesting to include a
toy dataset, and an unknown referee for many valuable remarks.

\noindent
{\sc Universit\"at Karlsruhe, Institut f\"ur Industrielle Bauproduktion, Englerstr.\ 7, 76131 Karlsruhe, Germany}

\noindent
Email: {\tt bradley@ifib.uni-karlsruhe.de}
\end{document}